\documentclass{article}

\usepackage{PRIMEarxiv}

\usepackage[utf8]{inputenc} 
\usepackage[T1]{fontenc}    
\usepackage{hyperref}       
\usepackage{url}            
\usepackage{booktabs}       
\usepackage{amsfonts}       
\usepackage{nicefrac}       
\usepackage{microtype}      
\usepackage{lipsum}
\usepackage{fancyhdr}       
\usepackage{graphicx}       
\graphicspath{{media/}}     

\usepackage{siunitx}
\usepackage{amsmath,amssymb}
\usepackage{algorithm}
\usepackage{algpseudocode}
\usepackage{multirow}

\title{Toward AI Autonomous Navigation for Mechanical Thrombectomy using Hierarchical Modular Multi-agent Reinforcement Learning (HM-MARL)
\thanks{\textit{\underline{Citation}}: 
\textbf{H. Robertshaw, N. Fischer, L. Karstensen, B. Jackson, X. Chen, SMH. Sadati,  C. Bergeles, A. Granados, and T. C. Booth (2026). Toward AI Autonomous Navigation for Mechanical Thrombectomy using Hierarchical Modular Multi-agent Reinforcement Learning (HM-MARL). IEEE Robotics and Automation Letters, doi: 10.1109/LRA.2026.3664661. }} 
}

\author{
  Harry Robertshaw, Nikola Fischer, Benjamin Jackson, Xingyu Chen, \\
  \textbf{Christos Bergeles, Alejandro Granados, Thomas C Booth†} \\
  School of Biomedical Engineering \& Imaging Sciences \\
  Kings College London \\
  London, UK\\
  \texttt{†thomas.booth@kcl.ac.uk} \\
   \And
  Lennart Karstensen \\
  AIBE \\
  Friedrich-Alexander University Erlangen-Nurnberg \\
  Erlangen, Germany\\
   \And
  S.M.Hadi Sadati \\
  School of Engineering \& Materials Science \\
  Queen Mary University of London \\
  London, UK\\
}

\begin{document}
\maketitle

\begin{abstract}
    Mechanical thrombectomy (MT) is typically the optimal treatment for acute ischemic stroke involving large vessel occlusions, but access is limited due to geographic and logistical barriers. Reinforcement learning (RL) shows promise in autonomous endovascular navigation, but generalization across `long' navigation tasks remains challenging. We propose a Hierarchical Modular Multi-Agent Reinforcement Learning (HM-MARL) framework for autonomous two-device navigation \textit{in vitro}, enabling efficient and generalizable navigation. HM-MARL was developed to autonomously navigate a guide catheter and guidewire from the femoral artery to the internal carotid artery (ICA). A modular multi-agent approach was used to decompose the complex navigation task into specialized subtasks, each trained using Soft Actor-Critic RL. The framework was validated in both \textit{in silico} and \textit{in vitro} testbeds to assess generalization and real-world feasibility. \textit{In silico}, a single-vasculature model achieved 92–100\% success rates on individual anatomies, while a multi-vasculature model achieved 56–80\% across multiple patient anatomies. \textit{In vitro}, both HM-MARL models successfully navigated 100\% of trials from the femoral artery to the right common carotid artery and 80\% to the right ICA but failed on the left-side vessel superhuman challenge due to the anatomy and catheter type used in navigation. This study presents the first demonstration of \textit{in vitro} autonomous navigation in MT vasculature. While HM-MARL enables generalization across anatomies, the simulation-to-real transition introduces challenges. Future work will refine RL strategies using world models and validate performance on unseen \textit{in vitro} data, advancing autonomous MT towards clinical translation.
\end{abstract}


\section{Introduction} \label{sec:introduction}

    Globally there were 93.8~million cases of stroke in 2021, with ischemic stroke accounting for 69.9~million of cases. From 2015 to 2035, the total direct medical stroke-related costs in the US are projected to more than double, from \$36.7~billion to \$94.3~billion~\cite{Martin2024}. Mechanical thrombectomy (MT) has emerged as a standard treatment for acute ischemic stroke resulting from large vessel occlusion, providing improved functional outcomes and lower mortality rates when compared with medical treatment alone~\cite{Bendszus2023, Nogueira2018, Goyal2016}. 
    
    Eligible patient outcomes are improved when MT is performed up to 24\,\unit{\hour} after stroke onset, however, a decreased interval from symptom onset to treatment has been shown to enhance therapeutic outcomes, with the benefit diminishing as time increases~\cite{Nogueira2018, Asdaghi2023}. In the UK, only 3.9\% of stroke admissions undergo MT despite $\approx 15\%$ of patients eligibility for treatment~\cite{SSNAP2024}, mainly because patients do not live within a MT-accessible catchment area. Alternatively, if patient transfer to a hospital providing MT is possible, long travel times may mean MT would be non-beneficial. Patients arriving from UK remote stroke centers are less likely to receive MT if the door-to-door time is $>$$3$\,\unit{\hour}, which occurs in 45\% of admissions~\cite{Zhang2021_2}. Similar findings have been seen in the US where the probability of undergoing MT decreases by 1\% for each additional minute of transfer time over an ideal transfer of 1\,\unit{\hour}~\cite{Regenhardt2018}. Other MT challenges include complications such as embolization of thrombus to a new brain territory, as well as perforation and dissection in the parent artery~\cite{Ngankou2021}. Operators also face X-ray radiation exposure, increasing cancer and cataract risks, while protective gear contributes to orthopedic strain \cite{Klein2009,Madder2017}.

    Robotic surgical systems offer a promising avenue to address these challenges through two complementary strategies, leading to an overall benefit of greater global MT accessibility due to the current unequal distribution of MT services~\cite{Robertshaw2023}. First, it is plausible that in endovascular specialties facing a shortage of highly-trained operators and MT centers, tele-operated MT may be performed safely and effectively by a few highly-trained operators based in centralized neuroscience centers. Second, less experienced operators, (e.g., interventional radiologists who are not used to performing neurointerventional procedures), may use assistive robots in peripheral hospitals with no neuroscience capabilities (there are only 27 centralized neuroscience centers in the UK, compared to two orders of magnitude more peripheral hospitals)~\cite{SSNAP2023}. It is also noteworthy that AI assistance could be used by highly trained operators in the first strategy to provide a `safety net' during tele-operated communication interruptions, as well as plausibly enhancing the procedural efficiency and safety of MT robots~\cite{RiveroMoreno2024}. For now, AI-enabled autonomy is envisioned as a complementary capability to tele-operation, rather than a direct replacement.

    Autonomous navigation of two devices has been investigated using reinforcement learning (RL) in two MT navigation tasks: 1) navigating a guidewire and catheter from the femoral artery to the internal carotid artery (ICA)~\cite{Robertshaw2024}, and 2) navigating a micro-catheter and micro-guidewire from the ICA to the middle carotid artery on unseen patient vasculatures~\cite{Robertshaw2025}. These studies were limited to \textit{in silico} work, and in the first, a single vasculature was used~\cite{Robertshaw2024}. RL autonomous single device navigation has been tested on \textit{in vitro} non-anatomical vessel platforms `idealized' for simple navigation, \textit{in vitro} aortic arches, and \textit{ex vivo} porcine livers~\cite{Cho2022, Chi2020, Karstensen2022}. However, these studies examine `short' navigations focused on a small anatomical section, as RL can struggle to adapt to complex tasks with `long' horizons (requiring multiple difficult inputs over a longer period)~\cite{Yu2019}. 
    
    A 2023 review found that there were no experimental testbeds assigned as benchmarks for comparing different RL algorithms~\cite{Robertshaw2023}. Since then, the open-source framework stEVE (simulated EndoVascular Environment) was released, which provides options for benchmarking RL algorithms~\cite{Karstensen2025,Moosa2025}. While stEVE has a \textit{DualDeviceNav} testbed benchmark for vasculature relevant to MT, experimentation was limited to \textit{in silico} testbeds, and no validation has yet been performed \textit{in vitro}. The development of \textit{in vitro} testbeds for MT - that integrate patient-specific vasculature and material properties - is crucial for bridging this gap and ensuring RL-trained policies translate effectively. However, training RL agents capable of generalizing to \textit{in vitro} testbeds relevant to MT requires training over diverse datasets reflecting normal variation in vasculature; as well as training over long navigation tasks requiring multiple device maneuvers (e.g., from the femoral artery to the ICA).

    Multi-Agent RL (MARL) presents a promising approach for addressing the challenges associated with long-horizon navigation tasks \textit{in vitro}~\cite{Zhang2021}. Current RL endovascular navigation studies train single agents, which can struggle with the sequential dependencies of multi-step navigation. Additionally, previous work on `long' navigation tasks has been limited to training and testing on a single \textit{in silico} vasculature, reducing generalizability and \textit{in vitro} translation~\cite{Robertshaw2024}. By training multiple agents, the navigation tasks can be broken up into smaller segments, allowing for more effective and computationally efficient training. In this study, we propose Hierarchical Modular Multi-Agent RL (HM-MARL), whereby the navigation task is split into sub-tasks with an associated specialized agent, while a hierarchical approach is used to determine which agent to use. Unlike prior MARL approaches that focus on concurrent agent collaboration, our method adopts a sequential, rule-based strategy, ensuring smooth transitions between navigation phases while maintaining efficiency and task modularity. This complements MT, which is sequential in nature and allows for factors such as imaging and device changes to take place as the procedure progresses.
    
    The aim of this study was to demonstrate \textit{in vitro} MT navigation from a femoral insertion point to a target in the ICA in realistic patient vasculature. The primary objective was to use HM-MARL to navigate two devices simultaneously from the femoral artery to the ICA. The secondary objectives were: 1) to develop a methodology for creating generalized, patient-specific MT environments using computed tomography angiography (CTA) scans, 2) to compare HM-MARL to the current state-of-the-art methods for autonomous MT navigation, and 3) to validate HM-MARL in both \textit{in silico} and \textit{in vitro} settings, assessing it’s generalizability across different vascular anatomies relevant for MT. Our contributions were as follows: 1) we introduce HM-MARL for autonomous endovascular navigation, demonstrating its ability to efficiently learn and execute complex, long-horizon tasks, which enabled us to 2) present the first \textit{in vitro} validation of RL-based MT navigation, and 3) present the first autonomous endovascular navigation using two devices \textit{in vitro}; furthermore, we 4) propose a scalable pipeline for generating diverse and anatomically accurate MT environments using CTA-derived vascular models, facilitating future research in RL-based surgical navigation. 

\section{Methods}

    \subsection{Navigation tasks}

\begin{figure}[]
            \centering
            \includegraphics[width=0.5\linewidth]{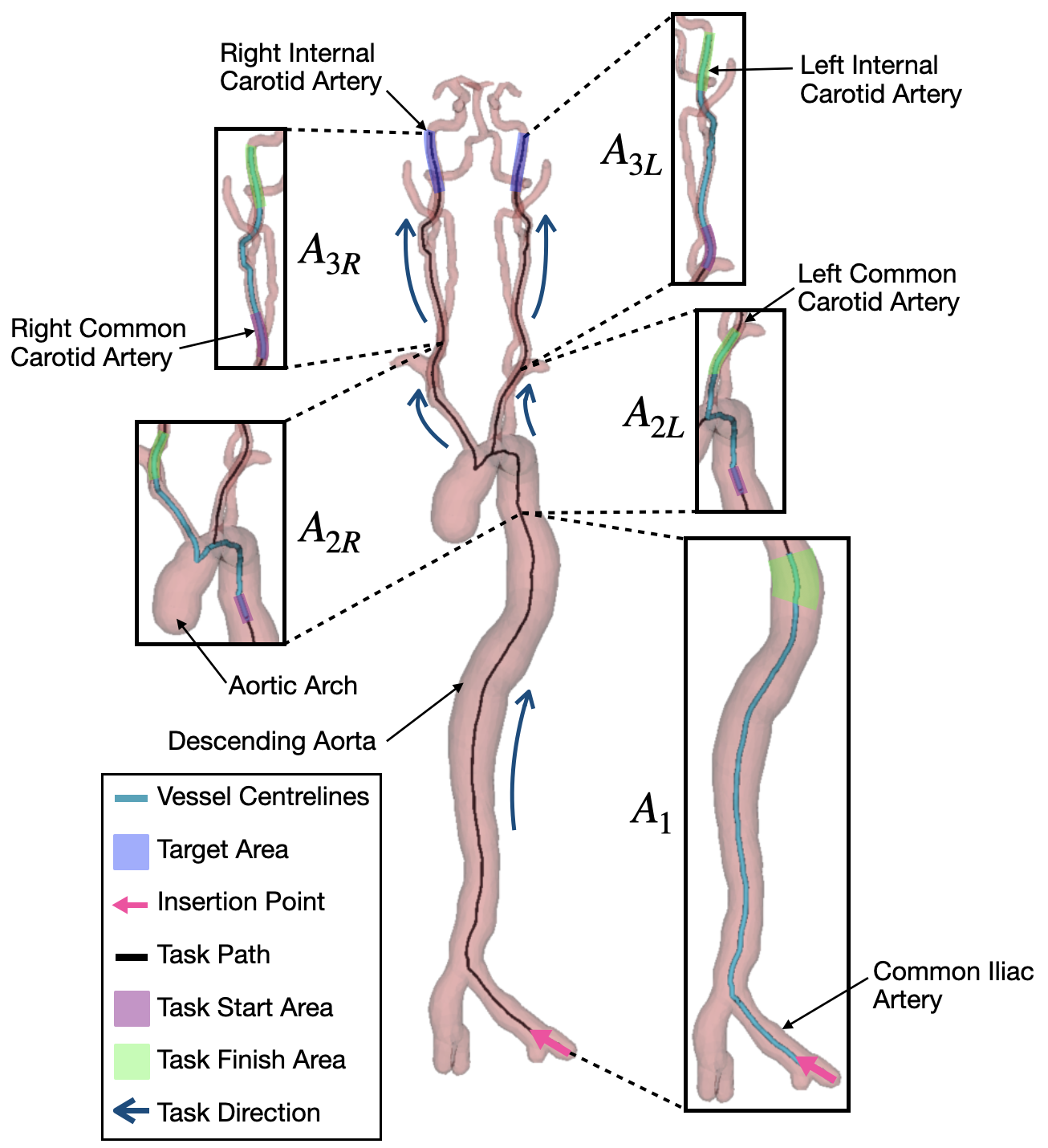}
            \caption{Full patient MT vasculature with sub-tasks labeled for the first phase of MT (navigating a guide wire and catheter from the common iliac artery the left or right ICA). Labels based on radiological left and right, where left sided anatomy is harder to navigate due to more acute angle of left common carotid artery (CCA) compared to the right. $A_1$ begins at the insertion point, hence no start area is labeled.}
            \label{fig:rober1}
        \end{figure}

        The MT navigation task has been split into five sub-tasks for training (Fig.~\ref{fig:rober1}), based on previously defined phases of MT~\cite{Crossley2019}. A target location and starting position were randomly sampled from a set of 20 points within the target vessel and starting vessel, respectively. For each navigation attempt, the target location (a sphere located along the centerline, with radius equal to that of the vessel cross-section), and the vasculature were changed.

    \subsection{Environment}

        \subsubsection{in silico}

            The stEVE framework was used to build an \textit{in silico} environment~\cite{Karstensen2025}. The BeamAdapter plugin for Simulation Open Framework Architecture (SOFA, v23.06) was used to model a Navien 0.058" support catheter (Medtronic, Minnesota, USA) (160 vertices with Young's modulus of 47\,\unit{\mega\pascal}) and 0.035" guidewire (Terumo, Tokyo, Japan) (120 vertices with Young's modulus of 43\,\unit{\mega\pascal})~\cite{Faure2012}. 
    
            A `neck CTA' scan that encompassed the aortic arch to the cerebral vessels, and a `body CTA' scan encompassing the abdominal and thoracic regions, including the femoral arteries, descending aorta, and the aortic arch (obtained with UK Research Ethics 24/LO/0057), were manually processed into surface meshes using 3D Slicer (v5.8.0)~\cite{Fedorov2012}. For all scans, the arterial centerlines and radii were then extracted. To fuse the neck and body CTAs, the centerlines of the abdominal and thoracic regions were scaled so that the radius at the most superior aspect of the descending aorta would match the neck CTA aortic arch centerline. These two sets of centerlines were joined, and the radius at each centerline point was used to generate a surface mesh, which was loaded into SOFA. This process was repeated for ten randomly selected neck CTAs, with the same body CTA scaled to fit each one. The mean right and left-hand-side vessel tortuosity were $1.19\pm0.07$~\unit{\milli\meter} and $1.14\pm0.05$~\unit{\milli\meter}, respectively, where the closer the value is to 1, the less tortuous it is, while Type-I aortic arches were found in 80\% (8/10) of vasculatures (the remaining were Type-II)~\cite{Lahlouh2023}. 
    
            Vasculature was augmented by random $\pm\,10$\unit{\degree} rotations and scaling (0.7 to 1.3 for height and width) to increase downstream generalizability, allowing domain adaptation for subsequent \textit{in vitro} experiments. A single-tracking (guidewire tracking only) method was employed~\cite{Robertshaw2024}. The simulation's fidelity to real-world guidewire behavior was achieved through iterative fine-tuning of parameters such as the friction between the guidewire and vessel wall and the guidewire's stiffness. Such methodology has previously allowed domain adaption of autonomous navigation agents from \textit{in silico} to \textit{in vitro} aortic arch models and from \textit{in silico} to \textit{ex vivo} of porcine liver vasculatures~\cite{Karstensen2025, Karstensen2022}.

        \subsubsection{in vitro}

\begin{figure}[]
                \centering
                \includegraphics[width=0.45\linewidth]{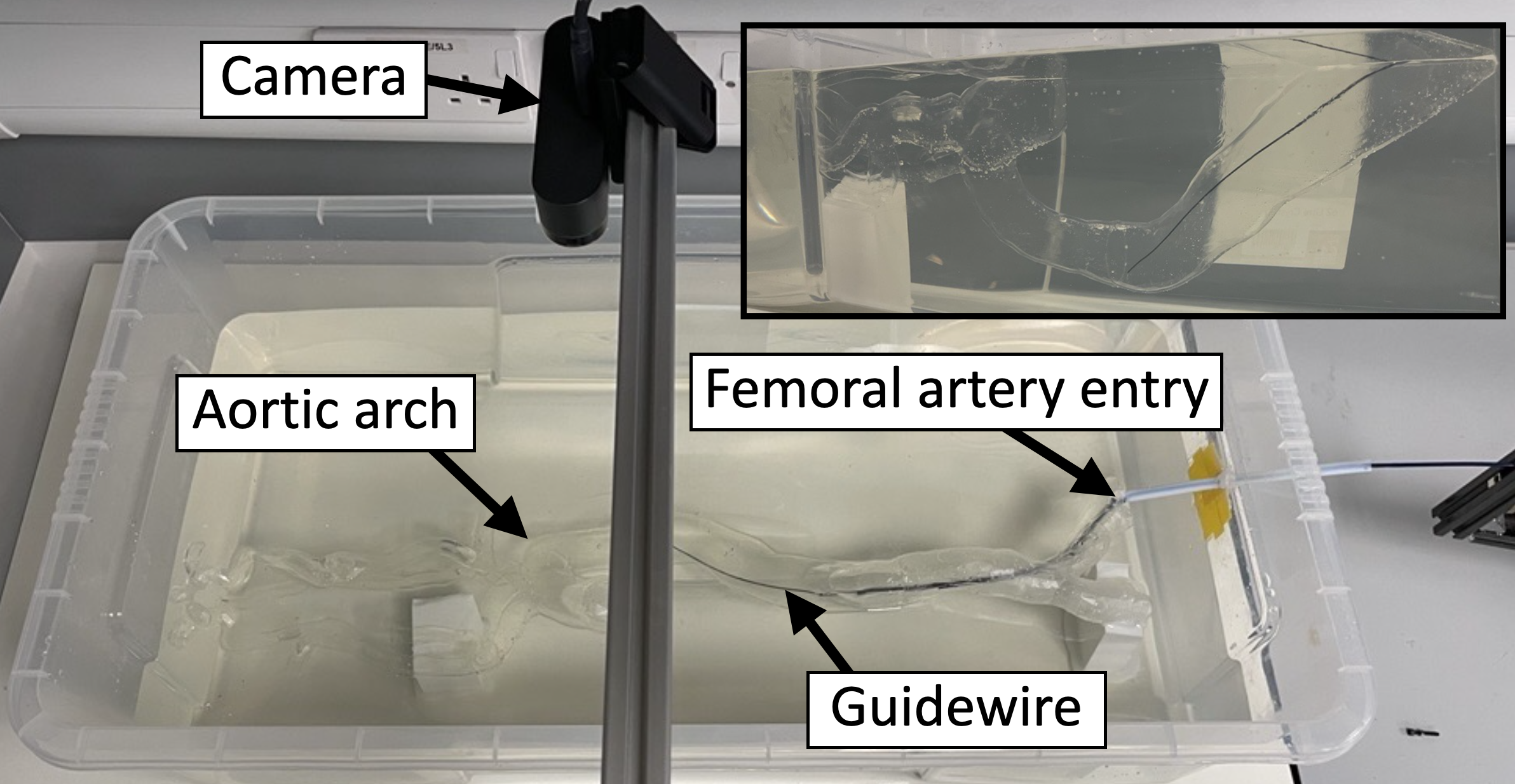}
                \caption{\textit{in vitro} testbed showing the 3D phantom and tracking camera. Inset is a side view of the 3D phantom.}
                \label{fig:rober2}
            \end{figure}

            Our \textit{in vitro} testbed (Fig.~\ref{fig:rober2}) consisted of a 3D-printed transparent vessel phantom (Clear V4, Formlabs Inc., Somerville, USA). The phantom was submerged in glycerine to improve the visibility of the device inside the phantom due to the similarity of its refractive index to that of the resin. A 4K camera (Logitech, Lausanne, Switzerland) mounted overhead was used to provide input to a device tip-tracking algorithm that extracted the necessary coordinates~\cite{Eyberg2022}.
 
            \begin{figure}[]
                \centering
                \includegraphics[width=0.45\linewidth]{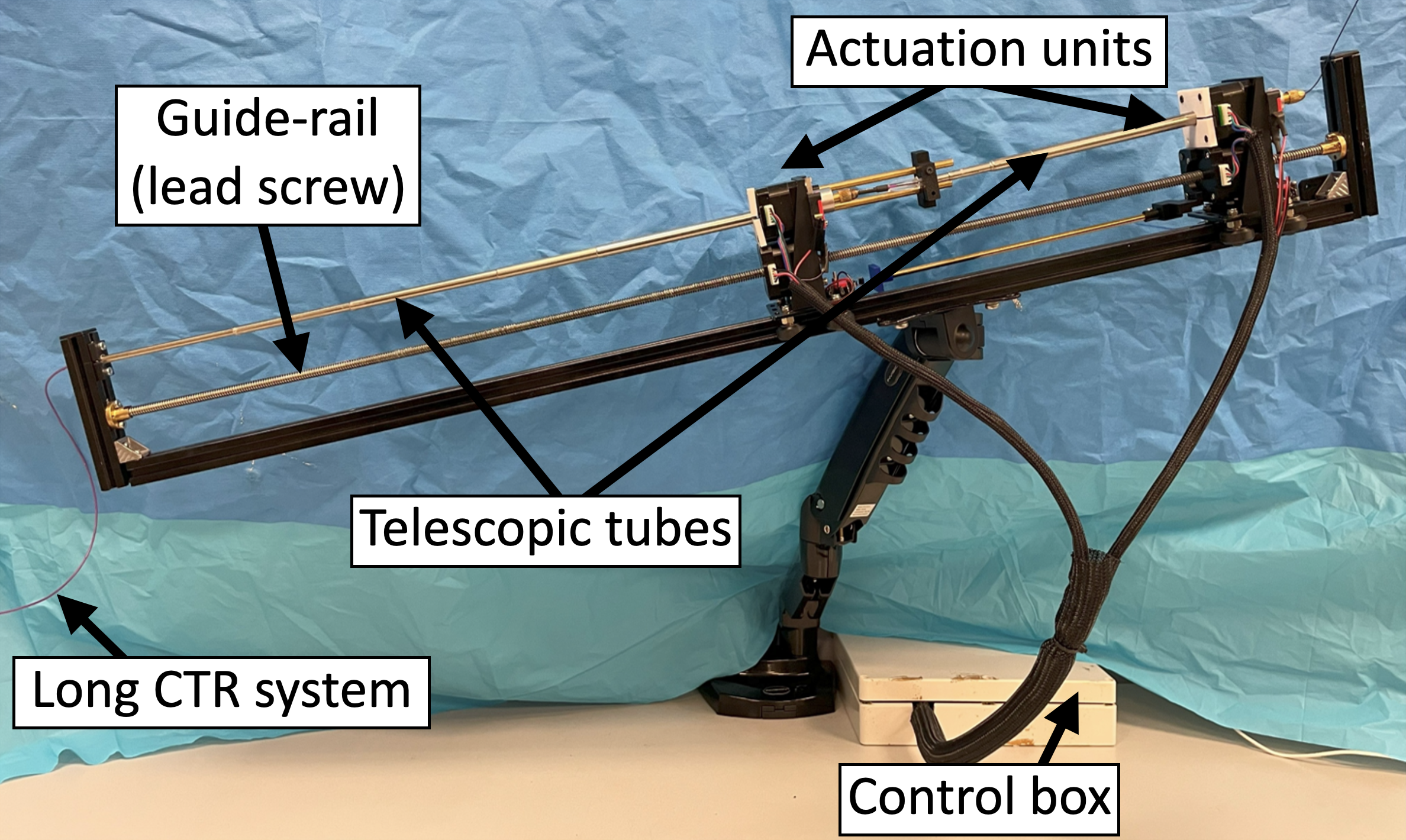}
                \caption{Concentric tube robot (CTR) manipulator used for \textit{in vitro} experiments~\cite{Sadati2025}.}
                \label{fig:rober3}
            \end{figure}

            A robotic manipulator (Fig.~\ref{fig:rober3}) was used to translate and rotate both the catheter and guidewire independently. The system featured a modular design with independent actuation carriage units, each capable of two degrees of freedom (rotation and translation), placed on a single lead screw rod (1\,\unit{\meter} long, 8\,\unit{\milli\meter} outer diameter, 2\,\unit{\milli\meter} pitch) which was fixed on a frame made from aluminum struts ($20 \times 20$\,\unit{\milli\meter} profile). Direct-drive hollow shaft stepper motors (NEMA 17, 1.68\,\unit{\ampere}) were used with a leadscrew nut to translate the actuation carriage unit with an instrument along the length of the actuation system. Furthermore, the same stepper motor type - but with a hollow shaft configuration - was deployed to rotate the instruments via small drill chuck couplings. Telescopic tubes were installed between the carriages to prevent lateral buckling of the catheters and guidewires. The motors were controlled via G-code protocol using a standard CNC control board (BIGTREETECH Octopus v1.1 control board). A 180\,\unit{\centi\metre} 0.035" guidewire (Terumo, Tokyo, Japan), and a 125\,\unit{\centi\metre} Navien 0.058" support catheter (Medtronic, Minnesota, USA) were used for experiments. The catheter does not have a shaped tip but it remains possible for an expert operator to navigate into the right common carotid artery (CCA). In contrast, when a straight catheter is combined with the acute angle of the left CCA, an expert operator cannot navigate the testbed – this allowed us to evaluate a superhuman additional navigation task (Fig.~\ref{fig:rober1}).

        \subsubsection{Agent's action space and observations}

            The device's distal position was described by three 2D coordinates equally spaced 2\,\unit{\milli\meter} apart along the guidewire, denoted as $(x_g', y_g')_{i=1,2,3}$, with $(x_g', y_g')_{1}$ representing the instrument tip; no visual information showing the geometry of the patient vasculature was given. Using this method, it is plausible that device tip points could be extracted from live fluoroscopic images~\cite{Eyberg2022} during MT, providing the inputs needed for the RL agent to navigate the vasculature in 3D. The target location was specified by the coordinates $(x_t', y_t')$. Observations comprised current and previous device positions, target location, and the previous action taken. The agent's action space included guidewire rotation and translation speed (both applied at the proximal device end). The rotational and translational speed were constrained to a maximum of 180\,\unit{\degree\per\second} and 40\,\unit{\mm\per\second}, respectively. 

    \subsection{Controller architecture}

        \subsubsection{Soft actor-critic}

            All RL models in this study were trained using the model-free RL algorithm, Soft Actor-Critic (SAC), whereby the critic learns the value and the actor optimizes the critic directly to maximize cumulative rewards. The implementation of this RL algorithm is the current state-of-the-art for autonomous endovascular interventions~\cite{Moosa2025,Karstensen2025}. The architecture includes a Long Short-Term Memory layer for learning trajectory-dependent state representations and feedforward layers for controlling the devices. The controller takes observations as input, and a Gaussian policy network outputs mean ($\mu$) and standard deviation ($\sigma$) for expected actions, representing the rotation and translation of the device (i.e., the catheter or guidewire). During training, actions are sampled from the $\sigma$, but for evaluation, $\mu$ is used directly for deterministic behavior. 
            
            A dense reward function was used during training, as shown in Equation~\ref{eq:R}~\cite{Robertshaw2023}. \textit{Pathlength} is defined as the centerline distance between the guidewire tip and the target, with $\Delta\text{pathlength}$ representing the change in pathlength from the previous step. Coefficients were chosen based on preliminary experiments to balance reward components and training stability. These values have shown positive outcomes for similar stEVE-based autonomous navigation tasks~\cite{Robertshaw2_2025}.
    
            \begin{equation}
                R = -0.00015 - 0.001\cdot\Delta\text{pathlength}+\begin{cases}1 & \text{if target reached}\\0 & \text{else}\end{cases}
                \label{eq:R}
            \end{equation}

        \subsubsection{HM-MARL}

\begin{figure}[]
                \centering
                \includegraphics[width=0.5\linewidth]{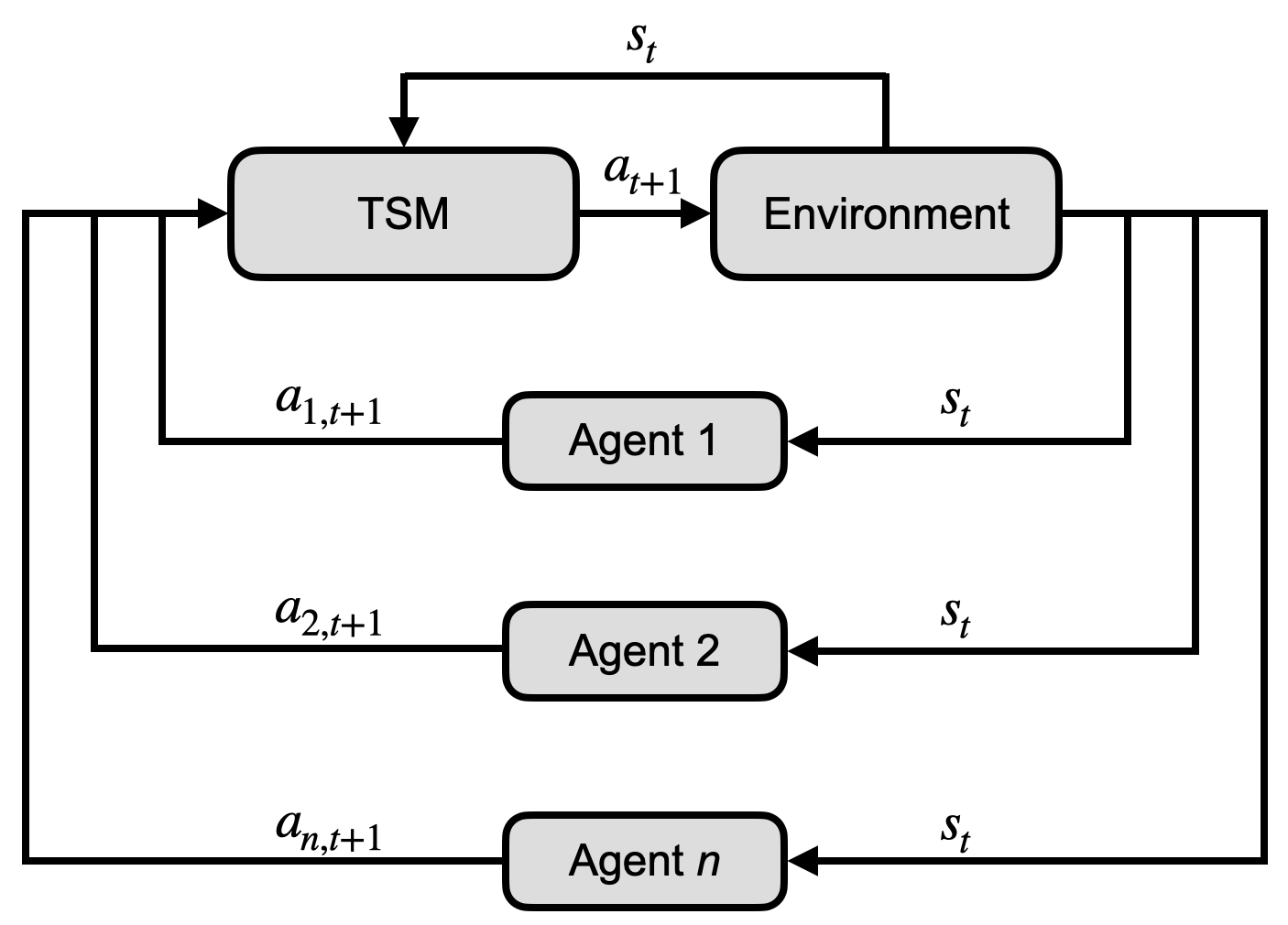}
                \caption{HM-MARL. TSM is used to select $a_{t+1}$ based on $s_t$.}
                \label{fig:rober4}
            \end{figure}

            MARL can be represented by a team Markov game, whereby agents collaborate to maximize a common reward, with each agent selecting an action simultaneously and the subsequent state depending on the joint actions~\cite{Wong2023}. While classical MARL assumes simultaneous multi-agent interaction, our approach adopts a modular multi-agent structure in which multiple specialized agents operate sequentially rather than concurrently. At every time step in HM-MARL (Fig.~\ref{fig:rober4}, Algorithm~\ref{alg:hm-marl}), each agent is given the current state, $s_{t}$, outputting an action for the next timestep, $a_{n,t+1}$ based on the pre-trained agent policy, $\pi_n$. These individual agents form the modular components of the system, each responsible for a distinct navigational sub-task. The task selection module (TSM) uses $a_{n,t+1}$ and $s_{t}$ to determine an appropriate final $a_{t+1}$ to give to the environment. The TSM functions as a higher-level controller, introducing a hierarchical structure by governing policy selection across sub-tasks without imposing classical hierarchical RL subgoal generation. The TSM's decision is based on the current position of the device and the target location (intermediate targets are used during the navigation).

            To utilize HM-MARL, five RL agents were trained (one for each defined sub-task in Fig.~\ref{fig:rober1}), each for $1 \times 10^7$ exploration steps, with a navigation task (or episode) considered complete when the target was reached within a threshold equal to the entire cross-section of the vessel. A timeout of 200 steps (approximately 27\,\unit{\second}) was set for training efficiency. Training was performed on an NVIDIA DGX A100 node (Santa Clara, California, USA) with 8 GPUs.

            \begin{algorithm}[]
                \footnotesize
                \caption{Hierarchical Modular Multi-Agent Reinforcement Learning (HM-MARL)}
                \label{alg:hm-marl}
                \begin{algorithmic}[1]
                    \State \textbf{Input:} Pre-trained policies $\pi_n$ (trained using SAC), TSM, environment
                    \State \textbf{Initialize:} Environment state $s_t$
                    \While{episode not terminated}
                        \For{each agent $n$ \textbf{in parallel}}
                            \State Receive global state $s_t$
                            \State Output action proposal $a_{n,t+1} \gets \pi_n(s_t)$
                        \EndFor
                        \State Aggregate proposed actions $\{a_{n,t+1}\}_{n=1}^N$
                        \State TSM selects final action $a_{t+1} \gets \text{TSM}(\{a_{n,t+1}\}, s_t)$
                        \State Execute action $a_{t+1}$ in the environment
                        \State Observe next state $s_{t+1}$
                        \State Update $s_t \gets s_{t+1}$
                    \EndWhile
                \end{algorithmic}
            \end{algorithm}

    \subsection{Experimental design and evaluation}\label{experiment_design}

        Three types of models were evaluated in this study:
        
        \begin{itemize}
            \item HM-MARL-1: A model trained and tested on one patient vasculature using HM-MARL - for proof-of-concept of HM-MARL.
            \item SA-RL-1: A model trained and tested on one patient vasculature using single agent RL - for comparison to the current state-of-the-art.
            \item HM-MARL-10: A model trained and tested on ten different patient vasculatures using HM-MARL - to show HM-MARL generalization.
        \end{itemize}

        \subsubsection{in silico}

            HM-MARL-1, HM-MARL-10 and SA-RL-1 were all evaluated \textit{in silico} on the same patient vasculature used in HM-MARL-1 and SA-RL-1 training. Additionally, HM-MARL-10 was also evaluated \textit{in silico} across the ten patient vasculatures used in its training. Long navigation tasks were created using concatenated shorter sub-tasks (Fig.~\ref{fig:rober1}). The following long navigation tasks were evaluated \textit{in silico}:
    
            \begin{itemize}
                \item $A_{1,2}$ (R) $= A_1 + A_{2R} \rightarrow$  Femoral to right CCA (RCCA).
                \item $A_{2,3}$ (R) $= A_{2R} + A_{3R} \rightarrow$ Descending aorta to right ICA (RICA).
                \item $A_{1,2,3}$ (R) $= A_1 + A_{2R} + A_{3R} \rightarrow$ Femoral to RICA.
                \item $A_{1,2}$ (L) $= A_1 + A_{2L} \rightarrow$ Femoral to left CCA (LCCA), superhuman task with straight catheter.
                \item $A_{2,3}$ (L) $= A_{2L} + A_{3L} \rightarrow$ Descending aorta to left ICA (LICA).
                \item $A_{1,2,3}$ (L) $= A_1 + A_{2L} + A_{3L} \rightarrow$ Femoral to LICA.
            \end{itemize}
    
            In all cases, 50\% of the target points in each branch were used in training, with the other 50\% reserved for evaluation, with the exact target points selected randomly and kept consistent across all models. Evaluations were conducted for 50~episodes, recording the success rate, procedure time, and path ratio. The \textit{Success Rate} (SR) was defined as the percentage of evaluation episodes in which the controller successfully reaches the target; \textit{Path Ratio} (PR) was defined as a measure of the remaining distance to the target point in unsuccessful episodes, calculated by dividing the remaining distance by the initial distance; and \textit{Procedure Time} (PT) was defined as the time taken from the start of navigation to the target location for successful episodes. The \textit{Exploration Steps} required to train each agent were recorded, defined as the number of training steps taken to reach the point at which the results are provided. A timeout of 1500~steps (200\,\unit{\second}) was used during \textit{in silico} evaluation due to the longer task length.

        \subsubsection{in vitro}

            HM-MARL-1 and HM-MARL-10 were then evaluated on the same \textit{in vitro} patient vasculature for long tasks ($A_{1,2}$ (R), $A_{1,2}$ (L), $A_{2,3}$ (R) and $A_{2,3}$ (L)). SA-RL-1 was omitted from \textit{in vitro} evaluation based on \textit{in silico} results. $A_{1,2,3}$ was not tested \textit{in vitro} due to a limited robotic manipulator stroke length. Evaluations \textit{in vitro} were conducted for 10~episodes, recording the SR, PT, and PR. A timeout of approximately 660\,\unit{\second} was used during \textit{in vitro} evaluation. Comparative statistical analyses were conducted using two-tailed paired Student's t-tests, with a predetermined significance threshold set at $p = 0.05$.

\section{Results}

    \subsection{in silico}

        \subsubsection{Single patient vasculature}

            Results from \textit{in silico} evaluation on a single patient vasculature can be seen in Table~\ref{tab:silico1}. For right-sided tasks ($A_{1,2}$ and $A_{2,3}$), HM-MARL-1 and HM-MARL-10 consistently achieved 100\% SR, while SA-RL-1 showed reduced SR across these tasks (e.g., 78\% for $A_{2,3}$ (R); $p < 0.001$). For left-sided tasks, the performance of SA-RL-1 dropped notably. On $A_{1,2}$ (L), it achieved only 18\% SR, which was worse than both HM-MARL variants ($p < 0.001$). HM-MARL-1 and HM-MARL-10 maintained near-perfect SR on this task, with substantially lower PTs (e.g., 44.0\,\si{\second} for HM-MARL-10; $p < 0.001$ vs.\ SA-RL-1). On $A_{1,2,3}$, SA-RL-1 failed to complete any trial (0\% SR). HM-MARL-1 maintained 100\% SR on the right-side version and 84\% on the left, significantly outperforming SA-RL-1 ($p < 0.001$). HM-MARL-10 also showed reasonable performance on this task (90\% on right, 60\% on left).
            
            \begin{table*}[]
                \centering
                \scriptsize
                \vspace{1ex}
                \caption{\textit{in silico} evaluation on a single patient vasculature, where each task was independent of each other. The reported values are mean $\pm$ standard deviation (standard deviation values may exceed logical bounds due to the statistical calculation). `$-$' denotes cell is not applicable (PT is not recorded when SR is 0\%, and PR is not recorded when SR is 100\%).}
                \label{tab:silico1}
                \begin{tabular}{c|ccc|ccc|ccc}
                    \multirow{2}{*}{\textbf{Task}} & \multicolumn{3}{c|}{\textbf{SA-RL-1}} & \multicolumn{3}{c|}{\textbf{HM-MARL-1}} & \multicolumn{3}{c}{\textbf{HM-MARL-10}} \\ \cline{2-10} 
                     & \multicolumn{1}{c|}{\textbf{SR (\%)}} & \multicolumn{1}{c|}{\textbf{PT (s)}} & \textbf{PR (\%)} & 
                       \multicolumn{1}{c|}{\textbf{SR (\%)}} & \multicolumn{1}{c|}{\textbf{PT (s)}} & \textbf{PR (\%)} &
                       \multicolumn{1}{c|}{\textbf{SR (\%)}} & \multicolumn{1}{c|}{\textbf{PT (s)}} & \textbf{PR (\%)} \\ \hline
                    {$A_{1,2}$ (R)} & \multicolumn{1}{c|}{$98 \pm 14$} & \multicolumn{1}{c|}{$33.2 \pm 18.3$} & $47 \pm 14$ & \multicolumn{1}{c|}{$100 \pm 0$} & \multicolumn{1}{c|}{$40.1 \pm 33.8$} & $-$ & \multicolumn{1}{c|}{$100 \pm 0$} & \multicolumn{1}{c|}{$77.8 \pm 51.4$} & $-$ \\ \hline
                    {$A_{2,3}$ (R)} & \multicolumn{1}{c|}{$78 \pm 42$} & \multicolumn{1}{c|}{$50.1 \pm 43.6$} & $22 \pm 13$ & \multicolumn{1}{c|}{$100 \pm 0$} & \multicolumn{1}{c|}{$22.3 \pm 9.9$} & $-$ & \multicolumn{1}{c|}{$100 \pm 0$} & \multicolumn{1}{c|}{$58.1 \pm 38.0$} & $-$ \\ \hline
                    {$A_{1,2,3}$ (R)} & \multicolumn{1}{c|}{$0 \pm 0$} & \multicolumn{1}{c|}{$-$} & $18 \pm 21$ & \multicolumn{1}{c|}{$100 \pm 0$} & \multicolumn{1}{c|}{$44.3 \pm 31.1$} & $-$ & \multicolumn{1}{c|}{$90 \pm 30$} & \multicolumn{1}{c|}{$53.8 \pm 28.3$} & $60 \pm 26$ \\ \hline
                    {$A_{1,2}$ (L)} & \multicolumn{1}{c|}{$18 \pm 39$} & \multicolumn{1}{c|}{$106.0 \pm 41.9$} & $66 \pm 21$ & \multicolumn{1}{c|}{$100 \pm 0$} & \multicolumn{1}{c|}{$45.9 \pm 29.6$} & $-$ & \multicolumn{1}{c|}{$100 \pm 0$} & \multicolumn{1}{c|}{$44.0 \pm 31.9$} & $-$ \\ \hline
                    {$A_{2,3}$ (L)} & \multicolumn{1}{c|}{$70 \pm 46$} & \multicolumn{1}{c|}{$37.5 \pm 28.4$} & $28 \pm 27$ & \multicolumn{1}{c|}{$94 \pm 24$} & \multicolumn{1}{c|}{$42.8 \pm 31.0$} & $70 \pm 16$ & \multicolumn{1}{c|}{$76 \pm 43$} & \multicolumn{1}{c|}{$35.2 \pm 23.1$} & $47 \pm 14$ \\ \hline
                    {$A_{1,2,3}$ (L)} & \multicolumn{1}{c|}{$0 \pm 0$} & \multicolumn{1}{c|}{$-$} & $20 \pm 28$ & \multicolumn{1}{c|}{$84 \pm 37$} & \multicolumn{1}{c|}{$60.1 \pm 23.0$} & $75 \pm 22$ & \multicolumn{1}{c|}{$60 \pm 49$} & \multicolumn{1}{c|}{$97.6 \pm 54.1$} & $62 \pm 8$ \\ 
                \end{tabular}
            \end{table*}

        \subsubsection{Multiple patient vasculatures}
        
            Evaluation across ten different patient vasculatures was conducted to investigate  if the model generalizes to a range of different vasculatures. Results can be seen in Table~\ref{tab:silico10}. On right-sided tasks for $A_{1,2}$ and $A_{2,3}$, SR remained high but was slightly reduced on multiple anatomies (98\% vs.\ 100\%, $p=0.315$; 92\% vs.\ 100\%, $p=0.039$). Correspondingly, PT increased non-significantly for $A_{1,2}$ (68.0\,\si{\second} vs.\ 77.8\,\si{\second}, $p=0.328$), and decreased for $A_{2,3}$ (33.1\,\si{\second} vs.\ 58.1\,\si{\second}, $p=0.001$). On left-sided tasks, however, the performance gap widened significantly. For $A_{1,2}$ (L), SR dropped from 100\% on a single vasculature to 62\% on multiple vasculatures ($p<0.001$). The lowest SR was for the longer $A_{1,2,3}$ task, where SR decreased from 90\% (single vasculature, right) to 66\% (multiple vasculatures, right) and from 60\% (single vasculature, left) to 46\% (multiple vasculatures, left) ($p=0.003$, $p=0.161$).
    
            \begin{table}[]
                \centering
                \scriptsize
                \caption{\textit{in silico} evaluation of the HM-MARL-10 model across ten different patient vasculatures, where each task was independent of each other.}
                \label{tab:silico10}
                \begin{tabular}{c|c|c|c}
                    \textbf{Task} & \textbf{\begin{tabular}[c]{@{}c@{}}SR (\%)\end{tabular}} & \textbf{\begin{tabular}[c]{@{}c@{}}PT (s)\end{tabular}} & \textbf{\begin{tabular}[c]{@{}c@{}}PR (\%)\end{tabular}} \\ \hline
                    $A_{1,2}$ (R) & $98 \pm 14$ & $68.0 \pm 48.3$ & \begin{tabular}[c]{@{}l@{}}$85 \pm 0$\end{tabular}\\ \hline
                    $A_{2,3}$ (R) & $92 \pm 27$ & $33.1 \pm 35.5$ & \begin{tabular}[c]{@{}l@{}}$74 \pm 12$\end{tabular}\\ \hline
                    $A_{1,2,3}$ (R) & $66 \pm 48$ & $61.1 \pm 41.2$ & \begin{tabular}[c]{@{}l@{}}$66 \pm 23$\end{tabular}\\ \hline
                    $A_{1,2}$ (L) & $62 \pm 49$ & $29.1 \pm 14.8$ & \begin{tabular}[c]{@{}l@{}}$69 \pm 11$\end{tabular}\\ \hline
                    $A_{2,3}$ (L) & $66 \pm 48$ & $43.5 \pm 32.3$ & \begin{tabular}[c]{@{}l@{}}$34 \pm 21$\end{tabular}\\ \hline
                    $A_{1,2,3}$ (L) & $46 \pm 50$ & $74.6 \pm 55.7$ & \begin{tabular}[c]{@{}l@{}}$57 \pm 16$\end{tabular}\\ 
                \end{tabular} 
            \end{table}

        \subsubsection{Exploration steps}
        
            Exploration steps required to train each task required for the HM-MARL and SA-RL models can be found in Table~\ref{tab:steps}. HM-MARL-1 achieves convergence with relatively few samples for simpler tasks like $A_1$ and $A_{2R}$, requiring only $0.75 \times 10^6$ steps each. In contrast, HM-MARL-10, which learns a policy across ten anatomies, requires more steps for generalization—up to $8.25 \times 10^6$ steps for $A_{2R}$. SA-RL-1 also demands more extensive exploration (e.g., $9.0 \times 10^6$ steps for $A_{1,2,3}$), meaning more time and computational resources are also needed during training.
    
            \begin{table}[]
                \centering
                \scriptsize
                \caption{Exploration steps required to train \textit{in silico} agents. $1 \times 10^6$ exploration steps $\approx 80$\,\unit{\minute}. `$-$' denotes cell is not applicable as training for a particular task was not required, due to the training procedure.}
                \label{tab:steps}
                \begin{tabular}{c|c|c|c}
                    \textbf{Task} & \textbf{\begin{tabular}[c]{@{}c@{}}SA-RL-1 Steps\end{tabular}} & \textbf{\begin{tabular}[c]{@{}c@{}}HM-MARL-1 Steps\end{tabular}} & \textbf{\begin{tabular}[c]{@{}c@{}}HM-MARL-10 Steps\end{tabular}} \\ \hline
                    $A_1$ & $-$ & $0.75 \times 10^6$ & $0.75 \times 10^6$ \\ \hline
                    $A_{2R}$ & $-$ & $0.75 \times 10^6$ & $8.25 \times 10^6$ \\ \hline
                    $A_{2L}$ & $-$ & $5.00 \times 10^6$ & $4.25 \times 10^6$ \\ \hline
                    $A_{3R}$ & $-$ & $1.50 \times 10^6$ & $6.75 \times 10^6$ \\ \hline
                    $A_{3L}$ & $-$ & $0.75 \times 10^6$ & $1.25 \times 10^6$ \\ \hline
                    $A_{1,2}$ & $4.5 \times 10^6$ & $-$ & $-$ \\ \hline
                    $A_{2,3}$ & $7.5 \times 10^6$ & $-$ & $-$ \\ \hline
                    $A_{1,2,3}$ & $9.0 \times 10^6$ & $-$ & $-$ \\
                \end{tabular}
            \end{table}

    \subsection{in vitro}

\begin{figure}[]
            \centering
            \includegraphics[width=0.45\linewidth]{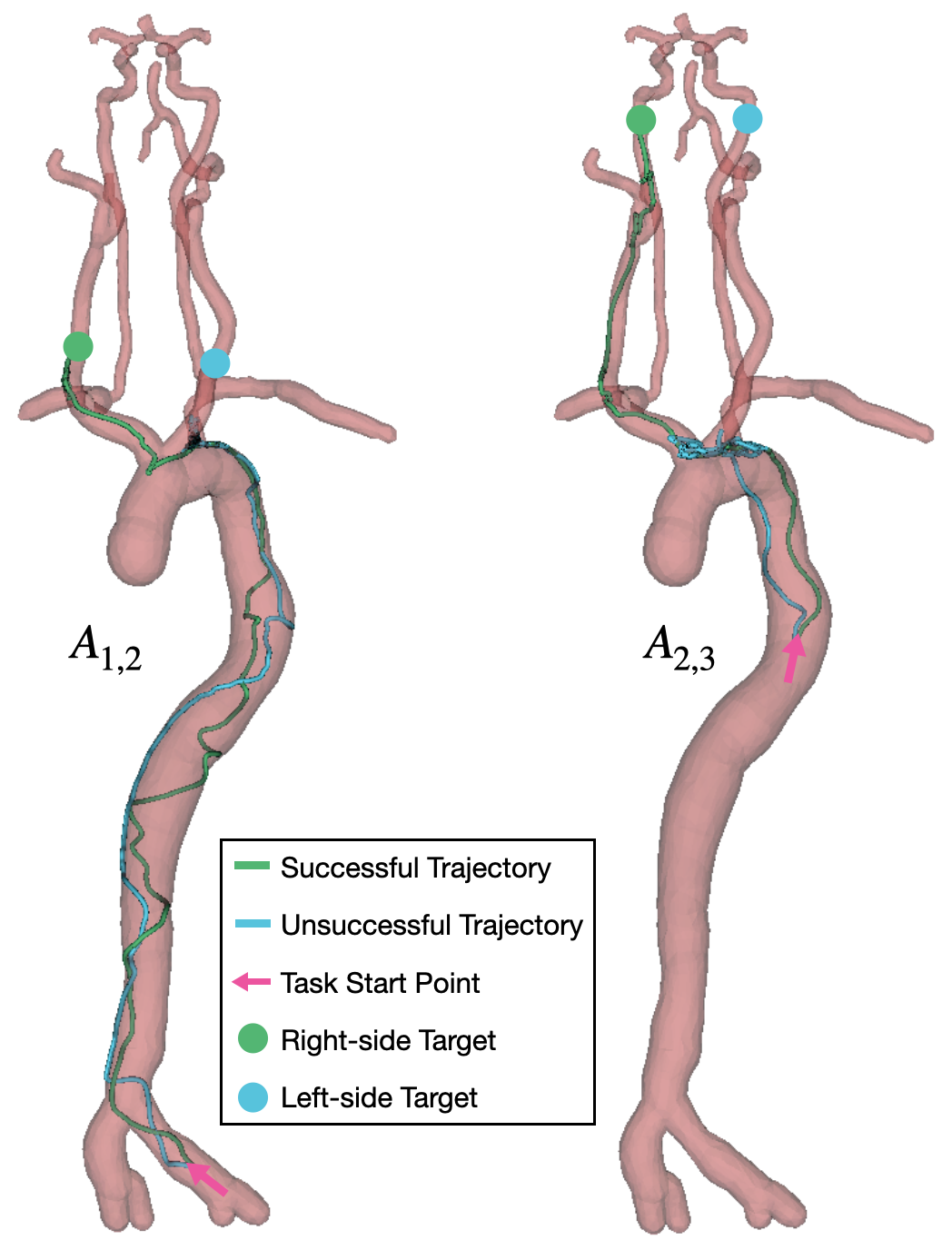}
            \caption{\textit{in vitro} navigation paths for $A_{1,2}$ and $A_{2,3}$. Successful trajectories are for right-sided navigation tasks, while unsuccessful trajectories are for left-sided navigation tasks.}
            \label{fig:rober5}
        \end{figure}

\begin{figure}[]
            \centering
            \includegraphics[width=0.9\linewidth]{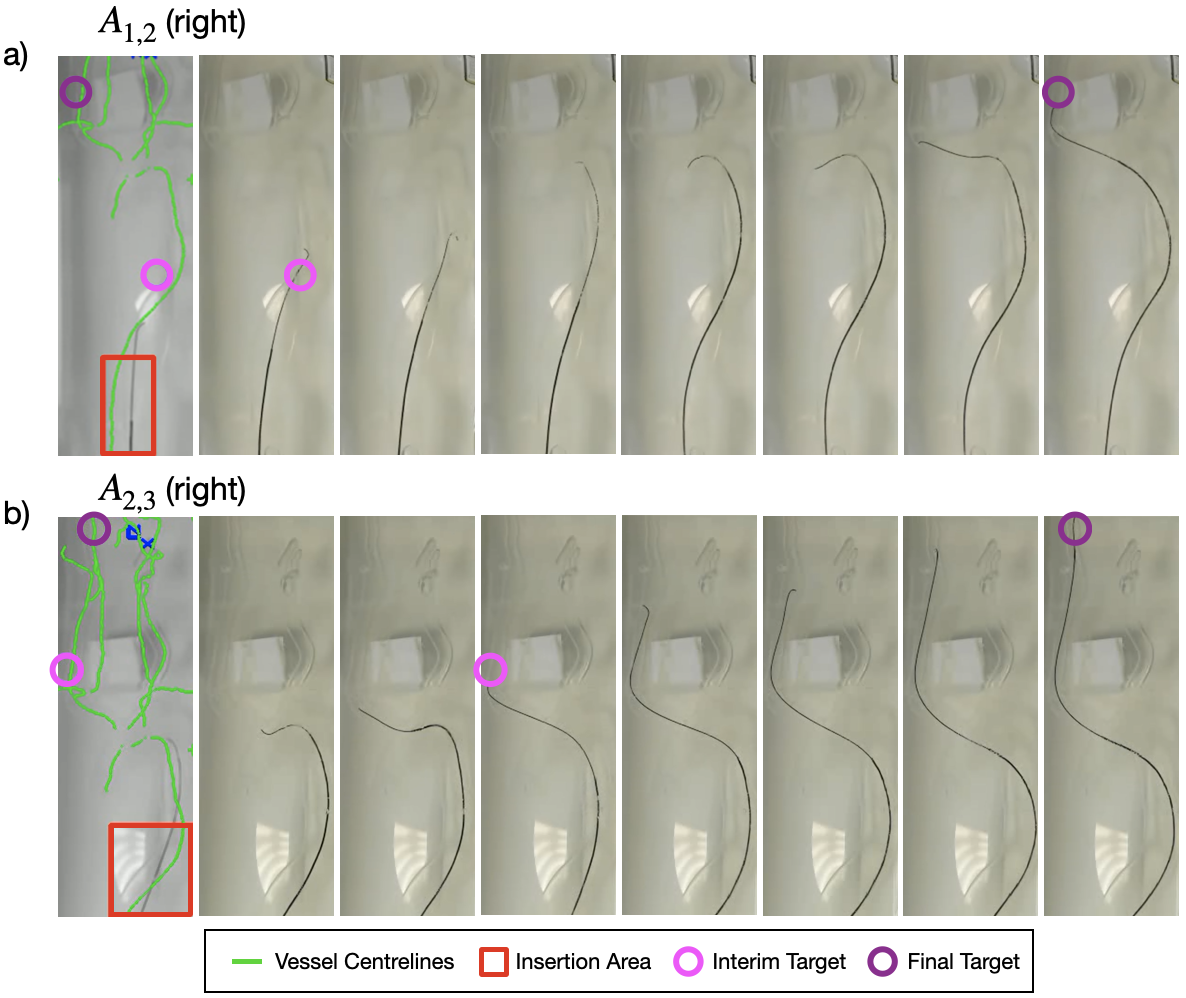}
            \caption{Selected frames from recording showing autonomous \textit{in vitro} navigation for a) $A_{1,2}$ (R), b) and $A_{2,3}$ (R).}
            \label{fig:rober6}
        \end{figure}

        Table~\ref{tab:vitro} shows results from \textit{in vitro} evaluation of HM-MARL-1 and HM-MARL-10. SA-RL-1 was omitted from \textit{in vitro} evaluation based on \textit{in silico} results. SRs of 100\% were recorded on task $A_{1,2}$ (R) for both models, while similar SRs were seen when comparing HM-MARL-1 to HM-MARL-10 for $A_{2,3}$ (R) (70\% to 80\% [$p=0.610$]). No successful navigations were recorded for either model on $A_{1,2}$ (L) or $A_{2,3}$ (L). There was no significant difference in PT or PR between the two models evaluated. The mean SR based on both left and right sides was 50\% for both models on $A_{1,2}$, 35\% for HM-MARL-1 on $A_{2,3}$, and 40\% for HM-MARL-10 on $A_{2,3}$. Fig.~\ref{fig:rober5} and \ref{fig:rober6} show example \textit{in vitro} navigation paths taken for all evaluated tasks. Whilst our intention was not to provide a comprehensive clinical comparison, an expert interventional neuroradiologist (10 years as UK consultant; US attending equivalent) achieved a $20$\% SR ($2/10$) on the left, and a $100$\% SR ($10/10$) on the right.

        \begin{table*}[]
            \centering
            \scriptsize
            \vspace{1ex}
            \caption{\textit{in vitro} evaluation of HM-MARL-1 and HM-MARL-10, where each task was independent of each other. `$-$' denotes cell is not applicable (PT is not recorded when SR is 0\%, and PR is not recorded when SR is 100\%).}
            \label{tab:vitro}
            \begin{tabular}{c|ccc|ccc}
                \multirow{2}{*}{\textbf{Task}} & \multicolumn{3}{c|}{\textbf{HM-MARL-1}} & \multicolumn{3}{c}{\textbf{HM-MARL-10}} \\ \cline{2-7} 
                 & \multicolumn{1}{c|}{\textbf{\begin{tabular}[c]{@{}c@{}}SR (\%)\end{tabular}}} & \multicolumn{1}{c|}{\textbf{\begin{tabular}[c]{@{}c@{}}PT (s)\end{tabular}}} & \textbf{\begin{tabular}[c]{@{}c@{}}PR(\%)\end{tabular}} & \multicolumn{1}{c|}{\textbf{\begin{tabular}[c]{@{}c@{}}SR (\%)\end{tabular}}} & \multicolumn{1}{c|}{\textbf{\begin{tabular}[c]{@{}c@{}}PT (s)\end{tabular}}} & \textbf{\begin{tabular}[c]{@{}c@{}}PR (\%)\end{tabular}} \\ \hline
                {$A_{1,2}$ (R)} & \multicolumn{1}{c|}{$100 \pm 0$} & \multicolumn{1}{c|}{$249 \pm 72$} & $-$ & \multicolumn{1}{c|}{$100 \pm 0$} & \multicolumn{1}{c|}{$261 \pm 62$} & $-$ \\ \hline
                {$A_{2,3}$ (R)} & \multicolumn{1}{c|}{$70 \pm 46$} & \multicolumn{1}{c|}{$421 \pm 67$} & $81 \pm 4$ & \multicolumn{1}{c|}{$80 \pm 40$} & \multicolumn{1}{c|}{$386 \pm 68$} & $87 \pm 7$ \\ \hline
                {$A_{1,2}$ (L)} & \multicolumn{1}{c|}{$0 \pm 0$} & \multicolumn{1}{c|}{$-$} & $73 \pm 4$ & \multicolumn{1}{c|}{$0 \pm 0$} & \multicolumn{1}{c|}{$-$} & $75 \pm 3$ \\ \hline
                {$A_{2,3}$ (L)} & \multicolumn{1}{c|}{$0 \pm 0$} & \multicolumn{1}{c|}{$-$} & $32 \pm 11$ & \multicolumn{1}{c|}{$0 \pm 0$} & \multicolumn{1}{c|}{$-$} & $36 \pm 7$ \\ 
            \end{tabular}
        \end{table*}

\section{Discussion}

    This study was the first to achieve \textit{in vitro} autonomous navigation in the MT vasculature. It showed effective navigation of the RCCA and the RICA from the femoral artery and the descending aorta, respectively. This represents the longest \textit{in vitro} autonomous endovascular navigation task completed. It was also the first endovascular \textit{in vitro} navigation which utilized two autonomous devices while navigating complex 3D patient vasculature. The proposed HM-MARL showed effective navigation over long tasks and across multiple vasculatures \textit{in silico}, effective \textit{in vitro} translation, and improvements over traditional SA-RL methods. These results provide an initial step toward advancing the technology readiness level (TRL) of current autonomous MT navigation systems toward level 4~\cite{Mankins1995}, principally by demonstrating right-sided validation.

    \subsection{in silico}

        The proposed HM-MARL method showed that it is possible to train separate agents for short RL tasks and combine them successfully to complete longer tasks during inference. While training and evaluating on a single patient vasculature showed promising SRs across all tasks, there was a decrease in performance observed when moving from HM-MARL-1 to HM-MARL-10 in $A_{2,3}$ (L) (94\% to 76\% [$p=0.011$]) and $A_{1,2,3}$ (right: 100\% to 90\% [$p=0.020$], left: 84\% to 60\% [$p=0.321$]). This decrease is due to a bias-variance trade-off, as HM-MARL-10 is able to navigate diverse vasculatures it sacrifices a slight performance degradation. The reduction in SR can also be attributed to the increased difficulty of moving from a test set of one vasculature to a test set of ten, with differing tortuosity, vessel radii and take-off. The aim of this proof-of-concept study was not to demonstrate complete generalization, but rather to show that the proposed framework can scale beyond a single anatomy (i.e., that the agent is not simply fitted to one). Furthermore, while SRs of 66\% (R) and 46\% (L - a superhuman experiment) for the HM-MARL-10 in task $A_{1,2,3}$ leaves room for improvement, it represents the best performing \textit{in silico} model for the first phase of MT navigation across multiple patient vasculatures~\cite{Karstensen2025,Moosa2025}.

        Additionally, the PR observed for HM-MARL-10 when evaluating across multiple vasculatures on $A_{1,2}$ (right: 68\%, left: 29\%), $A_{2,3}$ (right: 33\%, left: 44\%), and $A_{1,2,3}$ (right: 61\%, left: 75\%) indicates that the failure point in each case was the inability to perform task $A_{2}$. By finding improvements within this task alone, it would be possible to also increase the SR of the longer navigation tasks. Performance could potentially be improved by through using world models, a learned representation of the environment that simulates its dynamics, enabling a single agent to optimize actions in a virtual setting without relying solely on real-world data~\cite{Ha2018}. Recent work has shown that single configurations of RL algorithms (DreamerV3 and TD-MPC2) with no hyperparameter tuning can outperform specialized methods across diverse benchmark tasks~\cite{Hafner2023,Hansen2024}. This emerging area of research should be explored in the context of autonomous endovascular interventions.

    \subsection{in vitro}

        The first \textit{in vitro} autonomous navigation in the MT vasculature using an \textit{in silico} trained agent was shown in this study, with both models achieving a 100\% SR when navigating a guidewire and guide catheter from the femoral artery to the RCCA ($A_{1,2}$ (R)). Additionally, HM-MARL-10 recorded an 80\% SR when navigating a guidewire and guide catheter from the descending aorta to the RICA ($A_{2,3}$ (R)). While there was no significance between the HM-MARL-1 and HM-MARL-10 for $A_{2,3}$ (R) (70\% to 80\% [$p=0.610$]), it is hoped that HM-MARL-10 would generalize to different anatomies, making it able to adapt effectively to environmental differences from \textit{in silico} to \textit{in vitro}.

        A significant increase in PT was observed when moving from \textit{in silico} to \textit{in vitro} for both HM-MARL-1 ($A_{1,2}$: 40.1\,\unit{\second} to 249\,\unit{\second} [$p<0.001$], $A_{2,3}$: 22.3\,\unit{\second} to 421\,\unit{\second} [$p<0.001$]), and HM-MARL-10 ($A_{1,2}$: 77.8\,\unit{\second} to 261\,\unit{\second} [$p<0.001$], $A_{2,3}$: 58.1\,\unit{\second} to 386\,\unit{\second} [$p<0.001$]). This increase can be attributed to the step delay necessary for the robot to complete its movement. Inserting a small delay (0.5\,\unit{\second}) between each step ensured that the motors did not overheat, and allowed for the robot to fully complete the action before receiving the next one, at the detriment to total PT.

    \subsection{Implications for clinical translation and feasibility}

        The \textit{in vitro} results demonstrate that autonomous navigation can be learned, but fully autonomous RL-based control with no human intervention remains incompatible with current ethical, legal, and regulatory frameworks. The proposed method would use pre-trained, general agents for each task, which can overcome challenges associated with limited training times if used in current clinical pathways. For now, the most realistic route toward clinical translation is shared autonomy, whereby an autonomous agent assists with navigation, while the clinician retains decision-making authority, allowing for the agent to be stopped if necessary. Such systems would allow autonomous control to augment rather than replace human operators, providing a practical route for integrating RL-based navigation into MT.

    \subsection{Limitations and future work}

        While this study initiates progress toward TRL 4 for autonomous MT navigation systems (validation in laboratory environment), the current results represent partial laboratory validation; further evaluation (including broader anatomical coverage) is required for full TRL 4 classification. Deriving the tracking coordinates from fluoroscopy imaging as opposed to images from the top-down camera as in~\cite{Karstensen2025} would bring the experiments closer to the clinical environment. Additionally, the stroke length of the robotic manipulator meant that it was not possible to evaluate task $A_{1,2,3}$ (femoral artery to ICA) \textit{in vitro}. Future work should look to perform this task \textit{in vitro} by using a robotic manipulator capable of utilizing the entire length of the catheter and guidewire. 

        Despite the model's ability to navigate $A_{1,2}$ (L) and $A_{2,3}$ (L) \textit{in silico} with a minimum SR of 100\% and 76\% when evaluating on the same vasculature, neither of the trained models were able to perform left-sided navigation tasks \textit{in vitro}. Greater vessel tortuosity and unfavorable carotid artery take-off angles from the aortic arch have been shown to increase procedural difficulty~\cite{Shazeeb2023}. Techniques typically used to navigate the LCCA \textit{in silico} were not able to be replicated \textit{in vitro}, potentially due to differences in friction and device properties. An expert interventional neuroradiologist (10 years as UK consultant; US attending equivalent) had experimented on the testbed and confirmed that in a clinical scenario specialist catheters would be used, namely a guide catheter with a slightly bent tip, or a shaped slip catheter such as a Simmons 2 which has a reverse-curve. Whilst we had conducted a superhuman experiment on the left side, future experiments might test whether these instruments result in increased SRs for left-sided \textit{in vitro} navigation. Additionally, as in previous research~\cite{Chi2020}, a comparison of the proposed system to an expert operator must take place to determine how autonomous performance relates to clinical practice.

        While similar stEVE-based \textit{in silico} studies found that maximum forces reported were $\leq 1.0$\,\unit{\newton} (below a proposed vessel rupture threshold of 1.5\,\unit{\newton})~\cite{Robertshaw2025}, future work should look to measure forces exhibited on \textit{in vitro} vessel walls by using motor current as a proxy for axial or rotational load, ensuring that the agent doesn't exhibit any unsafe behaviors. The inclusion of vessel wall forces in the reward function should also be analyzed to determine if it can increase procedural safety \textit{in vitro})~\cite{Robertshaw2025}.

        This study evaluates a single anatomy \textit{in vitro} with no unseen data. While this work demonstrates promising results, the training, testing, and validation dataset must be expanded to include unseen data, and to replicate the different vasculature configurations or branching patterns of the found in the population. To enhance generalizability when expanding the dataset, future work could investigate a further breakdown of tasks to retain similar performance outcomes. Furthermore, the RL agents trained in this paper employed a state-based approach, which made use of a device tip tracking algorithm to provide an input to the agent \textit{in vitro}. Future work could compare this to an image-based RL method, whereby the entire image could be used as an input for the RL agent. Finally, future work may look to use a learning-based TSM, allowing it to adapt to unexpected dynamics such as delay, drift, or vessel deformation.

\section{Conclusions}

    This study represents the first demonstration of \textit{in vitro} autonomous navigation in the MT vasculature, marking a significant milestone in the development of robotic-assisted endovascular interventions. Our findings show that the proposed HM-MARL models achieve high SRs across long navigation tasks \textit{in silico}, and can generalize when trained and tested on multiple patient vasculatures. Our results indicate that while current HM-MARL agents demonstrate promising \textit{in vitro} navigation capabilities, the simulation-to-real transition introduces new challenges, including reduced SRs and increased sensitivity to environmental variations. Future work should focus on improving current RL agents using world models and moving toward unseen \textit{in vitro} validation. In summary, these results mark early progress in advancing the TRL of current autonomous MT navigation systems, indicating initial laboratory-level validation while highlighting the need for broader testing. Nonetheless, this work moves toward realizing the benefits of fully autonomous MT—and other endovascular procedures—in the clinic.

\section*{Acknowledgments}

    Partial financial support was received from the WELLCOME TRUST (Grant Agreement No 203148/A/16/Z), the Engineering and Physical Sciences Research Council Doctoral Training Partnership (Grant Agreement No EP/R513064/1), and the MRC IAA 2021 Kings College London (MR/X502923/1). For the purpose of Open Access, the Author has applied a CC BY public copyright license to any Author Accepted Manuscript version arising from this submission.

\bibliographystyle{unsrt}  
\bibliography{references}

\end{document}